\documentclass[lettersize,journal,twoside]{IEEEtran}
\usepackage{cite}
\usepackage{amsmath,amssymb,amsfonts}
\usepackage{algorithmic}
\usepackage{graphicx}
\usepackage{textcomp}
\usepackage{xcolor}
\usepackage{subfigure}
\usepackage{color,soul}
\usepackage{nomencl}
\usepackage{amsthm}
\usepackage{nicematrix}
\usepackage{comment}
\usepackage{multirow}
\usepackage{comment}
\usepackage{fancyhdr}
\newtheorem{theorem}{Theorem}

\begin{document}

\title{
Variable Impedance Control for Floating-Base Supernumerary Robotic Leg in Walking Assistance
}

\author{
Jun Huo, Kehan Xu, Chengyao Li, Yu Cao, Jie Zuo, Xinxing Chen$^*$ and Jian Huang$^*$,~\IEEEmembership{Senior Member,~IEEE}

\thanks{
Manuscript received: March 3, 2025; Revised June 2, 2025; Accepted July 4, 2025. 

This paper was recommended for publication by
Editor Pietro Valdastri upon evaluation of the Associate Editor and Reviewers'
comments.
This work was supported in part by the National Natural Science Foundation of China under Grant U24A20280, 62333007, and U1913207, and in part by the Program for HUST Academic Frontier Youth Team.

Jun Huo, Kehan Xu, Chengyao Li, Xinxing Chen and Jian Huang are with the Key Laboratory of the Ministry of Education for Image Processing and Intelligent Control and the Hubei Key Laboratory of Brain-inspired Intelligent Systems, School of Artificial Intelligence and Automation, Huazhong University of Science and Technology, Wuhan 430074, China.
Corresponding author: Xinxing Chen, \texttt{cxx@hust.edu.cn};
Jian Huang, \texttt{huang$\_$jan@mail.hust.edu.cn}.

Yu Cao is with the School of Electronic \& Electrical Engineering, University of Leeds, LS2 9JT Leeds, U.K. \texttt{y.cao1@leeds.ac.uk}.

Jie Zuo is with the School of Information Engineering, Wuhan University of
Technology, Wuhan 430070, China \texttt{zuojie@whut.edu.cn}.


Digital Object Identifier (DOI): see top of this page.
}}

\pagestyle{fancy}
\fancyhf{}
\renewcommand{\headrulewidth}{0pt}
\fancyhead{} 
\fancyhead[RE]{\scriptsize IEEE ROBOTICS AND AUTOMATION LETTERS. PREPRINT VERSION. ACCEPTED
JULY, 2025}
\fancyhead[LE]{\scriptsize \thepage}
\fancyhead[LO]{\footnotesize Huo $et$ $al$.: Variable Impedance Control for Floating-Base Supernumerary Robotic Leg in Walking Assistance}
\fancyhead[RO]{\scriptsize \thepage}
\markboth{IEEE ROBOTICS AND AUTOMATION LETTERS. PREPRINT VERSION. ACCEPTED
JULY, 2025}%
{Shell \MakeLowercase{\textit{et al.}}: A Sample Article Using IEEEtran.cls for IEEE Journals}

\maketitle


\begin{abstract}
In human-robot systems, ensuring safety during force control in the presence of both internal and external disturbances is crucial. As a typical loosely coupled floating-base robot system, the supernumerary robotic leg (SRL) system is particularly susceptible to strong internal disturbances.
To address the challenge posed by floating base, we investigated the dynamics model of the loosely coupled SRL and designed a hybrid position/force impedance controller to fit dynamic torque input.
An efficient variable impedance control (VIC) method is developed to enhance human-robot interaction, particularly in scenarios involving external force disturbances. 
By dynamically adjusting impedance parameters, VIC improves the dynamic switching between rigidity and flexibility, so that it can adapt to unknown environmental disturbances in different states. 
An efficient real-time stability guaranteed impedance parameters generating network is specifically designed for the proposed SRL, to achieve shock mitigation and high rigidity supporting. 
Simulations and experiments validate the system's effectiveness, demonstrating its ability to maintain smooth signal transitions in flexible states while providing strong support forces in rigid states.
This approach provides a practical solution for accommodating individual gait variations in interaction, and significantly advances the safety and adaptability of human-robot systems.
\end{abstract}


\section{Introduction}
Supernumerary robotic limbs (SRLs) can enhance the natural locomotor abilities of healthy individuals and assist in walking rehabilitation for hemiplegic patients \cite{yang2021supernumerary}.
As a floating-based robot system (FBRS), the SRL introduces additional complexity to its control due to the challenges associated with the floating base \cite{HamadTRO}.
At the same time, the unknown external contact environments will also cause certain interference to the control system \cite{LinTMECH}. 
These two kinds of internal and external disturbances will bring uncertainty and insecurity to the robot system.
Therefore, effectively managing internal disturbances introduced by the floating-base system, as well as external disturbances arising from interactions with the environment, are crucial for ensuring a safe and stable interaction.

Tightly coupled wearable exoskeletons, designed to align with human joints, can restrict natural movement and increase the risk of secondary injuries.
Such injuries may arise from joint misalignment, excessive motion, mechanical failure, and impaired balance, potentially leading to hyper-extension injuries, falls, skin irritation, or even long term musculoskeletal strain \cite{ Nasr2024,JunlinRAL}.
In contrast, SRL offers a viable alternative in wearable assistive robotics, representing a loosely coupled FBRS.
However, loose coupling introduces significant internal disturbances, making safe and stable force control more challenging. Traditional approaches model the floating base and the robotic system separately, lacking a parametric representation of coupling forces and failing to capture internal dynamic interactions.
Existing studies primarily focus on coordinating transformations \cite{Ref11} and constructing inverse dynamics \cite{Reher2021ICRA} to achieve effective system control.
For instance, Hyon et al. proposed a coordinate transformation technique to decouple joint-space and floating-base dynamics \cite{Ref13}, while Righetti et al. developed an inverse dynamics controller utilizing torque redundancy for optimal contact force distribution \cite{Ref15}. Henze et al. further integrated task hierarchies with passivity-based control for torque-controlled humanoids \cite{HenzeIJRR}, and Spyrakos et al. addressed flexible-joint balancing and locomotion for floating-base biped robots \cite{Spyrakos2023}.
However, the coordinating transformations method requires the observation of the interaction forces between the floating base and the SRL, and inverse dynamics requires the assumption that the contact points do not move relative to the inertial frame.
To address this gap, this study proposes a comprehensive dynamic model of the floating base–SRL system to estimate internal force disturbances. 
This formulation enables a parametric description of the coupling force interaction between the torso and the SRL.

To address the challenge of uncertain interaction disturbances during walking assistance, a variable impedance control (VIC) strategy is developed to enable the SRL system to switch adaptively between flexibility and rigidity. 
Unlike conventional approaches based on fixed impedance control, which may be suboptimal in unpredictable environments \cite{YTDong2019}, VIC allows dynamic adjustment of impedance parameters, emulating biological systems' intrinsic modulation mechanisms and enhancing adaptability in human–robot interaction \cite{Buchli2011}.
Specifically, when the SRL contacts the ground, it encounters impact forces that should be minimized to ensure user comfort and safety, necessitating low impedance for compliance. In contrast, during the support phase of walking, high impedance is required to provide sufficient assistive force. Thus, impedance should vary in response to the system's interaction state \cite{mizrahi2015mechanical}. 
Accordingly, designing a state-dependent VIC strategy capable of responding to unknown and time-varying conditions is essential for achieving robust and adaptive assistance.

When the FBRS makes contact with the ground, a closed kinematic chain is formed, introducing additional constraints and external disturbances to the system. In such scenarios, a state-dependent VIC strategy is particularly effective for ensuring stable and adaptive interaction.
Recent studies have demonstrated the advantages of VIC in various robotic applications \cite{He2020, XXzhang2020}.
Jin et al. reformulated VIC design as a control law optimization problem, enabling the incorporation of novel impedance constraints \cite{JinTMECH}. Pan et al. proposed a composite learning-based VIC strategy with three control modes under interval excitation, designed to achieve target impedance despite parametric uncertainties \cite{PanRAL}. Anand et al. introduced a deep model predictive VIC that dynamically adjusts the impedance parameters of a low-level VIC system \cite{anand2023model}. 
Despite these advancements, existing VIC approaches often neglect the explicit design of parameter update laws and fail to constrain the range of impedance variation, potentially compromising system stability \cite{Spyrakos2020}. A critical challenge lies in constructing an update law that ensures stability across the full range of impedance fluctuations \cite{Ficuciello2015}, given that system behavior is highly sensitive to the specific update dynamics.
One approach to address this issue involves constraining the variable impedance matrices to ensure stable impedance dynamics before task execution \cite{Kronander2016}. 
Under zero interaction force conditions, such constraints can ensure convergence of trajectory tracking errors \cite{He2020}.
In this paper, an efficient real-time stability guaranteed impedance parameters generating network (RSG-IPGN) in the VIC method is presented. The proposed method verifies the stability of impedance outputs in real time, thereby preventing instability that may arise from dynamic parameter updates.

SRLs for locomotor assistance are primarily configured as quadrupedal or tripedal systems. Quadrupedal SRLs are often inspired by quadruped animals and aim to enhance walking stability or support postures such as crawling and squatting \cite{PariettiTRO, Phillip, Fu2020JMR, zhang2024configuration}. However, this configuration usually results in higher mechanical complexity and greater physical demands on the user. In contrast, tripedal SRLs offer a lighter, more wearable solution for gait support with improved portability and adaptability \cite{Khazoom}.

Despite progress in mechanical design and trajectory generation, existing SRL studies mostly focus on gait pattern planning, with limited exploration of force control, especially under realistic multi-source disturbances during human-robot interaction. Unlike non-wearable assistive devices such as cane robots \cite{DiPeiTMech} and walkers \cite{HuangTMech}, which typically deal with external disturbances only, wearable SRLs must also manage internal perturbations arising from user movement and dynamic coupling.
To this end, our work presents a state-dependent VIC strategy uniquely tailored to the FBRS formed by the loosely coupled SRL and human user. The
main contributions of this article are listed as follows.

\begin{enumerate}
\item 
A model-based loosely coupled dynamic formulation of the human torso and SRL is developed to address internal disturbances induced by torso motions. A hybrid position/force control strategy is proposed, incorporating floating-base state feedback to enhance internal disturbance observation and compensation.

\item 
A state-dependent VIC strategy is proposed, and a real-time impedance generation network RSG-IPGN is proposed, enabling adaptive modulation of joint impedance while guaranteeing closed-loop stability.

\item A SRL prototype system is presented, and the proposed modeling and control method are verified through the human-SRL robot system, yielding comfortable and safety human robot interaction. 
\end{enumerate}

The rest of the paper is organized as follows:
Section II presents the VIC strategy for SRL, including loosely coupled dynamic modeling, and the real-time stability guaranteed impedance parameters generating network.
Section III describes the numerical simulations and section IV introduces the  human walking experiments.
Finally, conclusions are drawn in Section V.

\section{Methods}
\label{sec:guidelines}
\begin{figure*}[!t]    
\centering      
\includegraphics[width=0.85\textwidth]{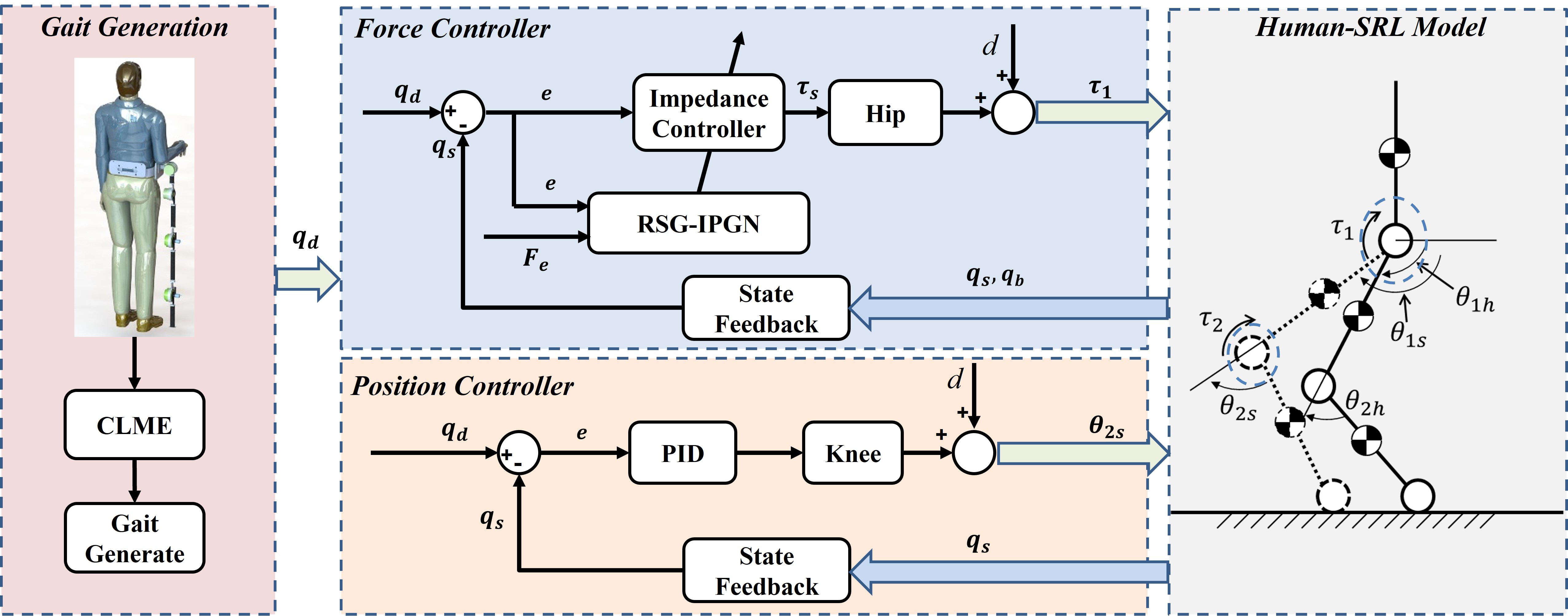}
\caption{The proposed FBRS-VIC real-time control system based on the Matlab/Simulink environment.}
\label{Overview}
\end{figure*}

In this section, the loosely coupling modeling of the  SRL is proposed. Then, a VIC framework with real-time stability guaranteed impedance parameters generating network is introduced. Fig.~\ref{Overview} summarizes the gait generation and the VIC controller.

The SRL system is a representative example of wearable FBRS, whose simplified profile is shown in Fig.~\ref{humanmodel}. 
The proposed SRL system has two degrees of freedom, which imitates the design of human hip and knee joints to assist the subject to walk.
The floating base, generated by human body movement, including the torso, head of human, is represented by the blue polygon. 
The manipulators, including human limbs (left arm, right arm, left leg and right leg) and SRL, attached in the floating base. 
Since SRL is a mechanism that can have independent gait and only one constraint is imposed by the human body on the SRL. This is due to the physical attachment between the two systems, called loosely coupling attachment.

\subsection{Gait Trajectory Generation}
The goal of SRL is to follow human movement and provide support, which requires generating a predetermined trajectory. This trajectory maps the user’s healthy leg movement to the SRL's motion, with the joint space trajectory of the healthy leg serving as the reference. We utilize inertial measurement units (IMUs) to measure the joint space trajectory of the leg.
Complementary limb motion estimation (CLME) is an effective method for estimating reference motion trajectories,  offering real-time performance, improved coordination, and strong adaptability, making it ideal for dynamic environments and gait planning in rehabilitation assistive systems.
The goal of CLME is a mapping function that outputs reference states (angles and velocities) for SRL directly in dependence of the current states of sound limbs \cite{CLME}.
\begin{equation}
\label{Gait1}
{\rm min}\quad  ||\boldsymbol{x_s} - \mathbf C \boldsymbol{x_{h}}||^2,
\end{equation}
where $\boldsymbol{x_h^T}=[\boldsymbol{\theta_h^T}, \boldsymbol{\dot \theta_h^T}]^T$ is the joint state vector of human leg, $\boldsymbol{x_{s}}=[\boldsymbol{\theta_s^T}, \boldsymbol{\dot \theta_s^T}]^T$ is the joint state vector of SRL. $\mathbf C$ is a mapping matrix which need to be calibrated. Then we can obtain the joint vector of SRL $\boldsymbol{x_{s}}$  as follows,
\begin{equation}
\label{Gait2}
\boldsymbol{\hat x_{s}} = \mathbf C \boldsymbol{x_h}.
\end{equation}

\subsection{Hybrid Position/Force Control based on Loosely Coupled Floating-base Dynamics Model}
\begin{figure}[tbp]    
\centering      
\includegraphics[width=1\columnwidth]{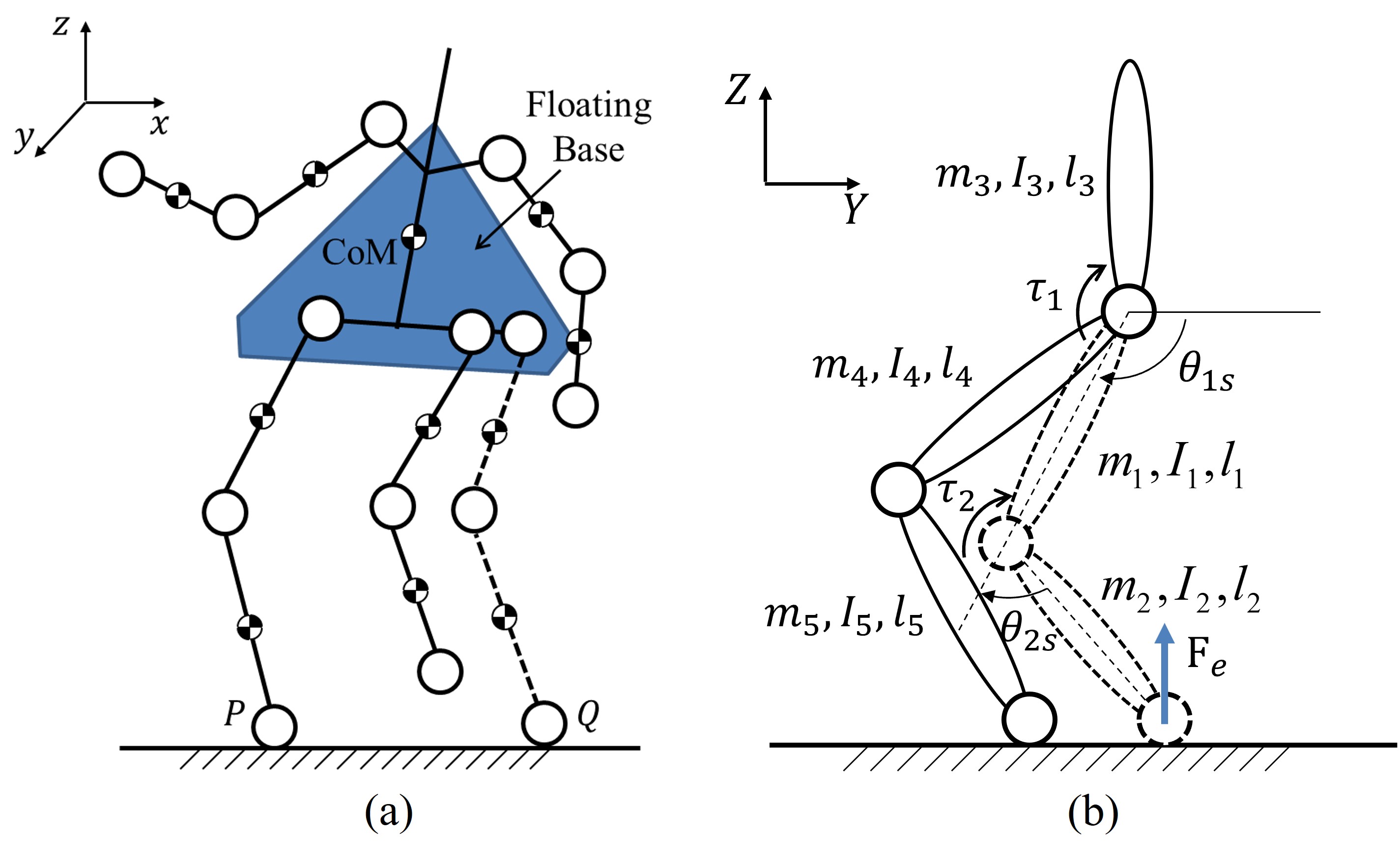}
\caption{The coordinate definition in coronal view (a) and sagittal view (b). The solid line represents the human body and the dashed line represents the SRL.}   
\label{humanmodel}
\end{figure}

The human-SRL system can be viewed as a human upper torso as a floating base, with human limbs and SRL mechanism regarded as end manipulator, shown in Fig.~\ref{humanmodel}.
The dynamics model of the whole body is as follows, the first row refers to the torso dynamics, while the second row represents the human limbs and SRL dynamics, which is,
\begin{equation}
\label{Dynamics}
\begin{aligned}
& \begin{bNiceArray}{cc}[cell-space-limits=2pt,margin=2pt]
\mathbf M_{bb}(\boldsymbol q) & \mathbf M_{bs}(\boldsymbol q)\\
\mathbf M_{bs}^T(\boldsymbol q) & \mathbf M_{ss}(\boldsymbol q)
\CodeAfter
\UnderBrace[yshift=3pt]{2-1}{2-2}{\mathbf{M}(\boldsymbol q)}
\end{bNiceArray}
\begin{bNiceArray}{cc}[cell-space-limits=2pt,margin=2pt]
\boldsymbol{\ddot q_b}\\
\boldsymbol{\ddot q_s} 
\CodeAfter
\UnderBrace[yshift=3pt]{2-1}{2-1}{\boldsymbol{\ddot q}}
\end{bNiceArray} 
+ \begin{bNiceArray}{cc}[cell-space-limits=2pt,margin=2pt]
 \mathbf C_{bb}\boldsymbol{(q,\dot q)} &\mathbf C_{bs}\boldsymbol{(q,\dot q)}\\
 \mathbf C_{sb}\boldsymbol{(q,\dot q)} &\mathbf  C_{ss}\boldsymbol{(q,\dot q)}  
\CodeAfter
\UnderBrace[yshift=3pt]{2-1}{2-2}{\mathbf C\boldsymbol{(q,\dot q)}}
\end{bNiceArray}\\ \\ \\
&\quad \begin{bNiceArray}{cc}[cell-space-limits=2pt,margin=2pt]
\boldsymbol{\dot q_b}\\
\boldsymbol{\dot q_s}
\CodeAfter
\UnderBrace[yshift=3pt]{2-1}{2-1}{\boldsymbol{\dot q}}
\end{bNiceArray} 
+ \begin{bNiceArray}{cc}[cell-space-limits=2pt,margin=2pt]
 \boldsymbol{\mathbf G_b(q_b)}\\
 \boldsymbol{\mathbf G_s(q_s)}   
\CodeAfter
\UnderBrace[yshift=3pt]{2-1}{2-1}{\boldsymbol{\mathbf G(q)}}
\end{bNiceArray} = \begin{bNiceArray}{cc}[cell-space-limits=2pt,margin=2pt]
\boldsymbol{\tau_b}\\
\boldsymbol{\tau_s}
\CodeAfter
\UnderBrace[yshift=6pt]{2-1}{2-1}{\boldsymbol{\tau}}
\end{bNiceArray} + \begin{bNiceArray}{cc}[cell-space-limits=2pt,margin=2pt]
  \boldsymbol{J_b^T(q_b)}\\
  \boldsymbol{J_s^T(q_s)}
\CodeAfter
\UnderBrace[yshift=3pt]{2-1}{2-1}{\boldsymbol{J^T}}
\end{bNiceArray}
\begin{bNiceArray}{cc}[cell-space-limits=2pt,margin=2pt]
\boldsymbol{F_h}\\
\boldsymbol{F_s}
\CodeAfter
\UnderBrace[yshift=6pt]{2-1}{2-1}{\boldsymbol{F_e}}
\end{bNiceArray},
\end{aligned}
\end{equation}
$\\$
where $\boldsymbol q=[\boldsymbol q_b^T, \boldsymbol q_s^T]^T$ is the generalized coordinates of the floating base and the SRL.
$\boldsymbol {\mathbf M(q)}$, $\boldsymbol {\mathbf C(q,\dot q)}$ and $\boldsymbol {\mathbf G(q)}$ are the mass matrix, coriolis matrix, and the gravitational vector, respectively.
$\boldsymbol \tau =[\boldsymbol{\tau_b, \tau_s}]^T$ is a vector of generalized torques actuated by motors. The matrix $\boldsymbol J=\partial \boldsymbol G / \partial \boldsymbol q$ represents the nonholonomic constraint matrix of the humanoid legs in contact, which is chosen as the position of the end-effector staying at ground level after impact. The force $\boldsymbol F_e=[\boldsymbol F_h, \boldsymbol F_s]^T$ is the external force, which is ground reaction force under our environmental conditions.

The purpose of the controller is to achieve hybrid force-position control, in which impedance control is applied to joint 1 to achieve variable impedance regulation adapting to the impact force, while achieving high stiffness support when contacting with ground. Joint 2 applies position control to  track the pre-defined trajectory.
Fig.~\ref{Overview} illustrated the diagram of variable impedance control strategy.

In dynamic model, the influence of floating base on SRL is considered as input interference. The disturbance due to environment is denoted as $\boldsymbol{\tau_d}$.
Correspondingly, the second row of the dynamic model Eq.~\eqref{Dynamics} can be rewritten to a nominal model,
\begin{equation}
\label{MB1}
\begin{aligned}
&\boldsymbol{\mathbf M_{ss}(q_s)\ddot q_s + \mathbf C_{ss}(q_s,\dot q_s)\dot q_s + \mathbf G_s(q_s)}\\
& \quad \quad + \boldsymbol{\mathbf M_{bs}^T(q_b)\ddot q_b} + \boldsymbol{\mathbf C_{sb}(q_b,\dot q_b)\dot q_b} \\
& \quad \quad = \boldsymbol{\tau_s} + \boldsymbol {J^T_s(q_s)F_s} + \boldsymbol{\tau_d}.
\end{aligned}
\end{equation}
where $\boldsymbol{q_s}=[\theta_{1s}, \theta_{2s}]^T$, $\boldsymbol{\tau_s}=[\tau_1, \tau_2]^T$, $\boldsymbol{F_s}=[0, F_e]^T$.

The objective of impedance control is to satisfy the following impedance model between position $\boldsymbol{q_s}$ and external force $\boldsymbol{F_s}$,
\begin{equation}
\label{MB2}
\mathbf M(\boldsymbol{\ddot q_s}-\boldsymbol{\ddot q_d})+\boldsymbol B(t)(\boldsymbol{\dot q_s}-\boldsymbol{\dot q_d})+\boldsymbol K(t)(\boldsymbol{q_s-q_d})=\boldsymbol{J_s^TF_s},
\end{equation}
where $\mathbf M, \boldsymbol B(t)$ and $\boldsymbol K(t)$ are impedance parameters;
$\boldsymbol B(t)$ and $\boldsymbol K(t)$ are diagonal matrix, which are $diag(B_1, B_2,..., B(t))$ and $diag(K_1, K_2,..., K(t))$. 
$\boldsymbol{q_d}$ is the desired position vector in joint space.
Following Eq.~\eqref{MB2}, yielding the vector $\boldsymbol{\ddot q_s}$,
\begin{equation}
\label{MB3}
\boldsymbol{\ddot q_s}=\mathbf M^{-1}(-\boldsymbol B(t)(\boldsymbol{\dot q_s}-\boldsymbol{\dot q_d})-\boldsymbol K(t)(\boldsymbol{q_s-q_d})+\boldsymbol{J_s^TF_s})+\boldsymbol{\ddot q_d},
\end{equation}
substituting Eq.~\eqref{MB3} into Eq.~\eqref{MB1}, the impedance controller can be designed to be
\begin{equation}
\label{MB4}
\begin{aligned}
&\boldsymbol \tau_s =  \underbrace{\boldsymbol{\mathbf M_{ss}\ddot q_d + \mathbf M_{ss}\mathbf M^{-1}B(t)\dot q_d + \mathbf M_{ss}\mathbf M^{-1}K(t)q_d}}_{\rm desired \, trajectory}\\
&  + \underbrace{\boldsymbol{(\mathbf C_{ss}-\mathbf M_{ss}\mathbf M^{-1}B(t))\dot q_s - \mathbf M_{ss}\mathbf M^{-1}K(t)q_s +\mathbf G_{s}}}_{\rm SRL \,state \,feedback}\\
&  + \underbrace{\boldsymbol{\mathbf M_{bs}^T\ddot q_b + \mathbf C_{sb}\dot q_b}}_{\rm torso\, state \, feedback}  + \underbrace{\boldsymbol{J_s^T(\mathbf M_{ss}\mathbf M^{-1}-I)F_s}}_{\rm force \,feedback}-\underbrace{\boldsymbol{\tau_d}}_{\rm dis}.
\end{aligned}
\end{equation}

As Eq.~\eqref{MB4}, the impedance controller input consists of three parts. The first part is related to the desired trajectory such that $\boldsymbol{q_d=x_s}$, which is known from Eq.~\eqref{Gait2}. The second part is related to the state of the current system. Because of the coupling of the floating system, there are two kinds of state feedback, SRL state feedback and torso state feedback. The last part is the force feedback. 
To eliminate the force feedback item, we can select $\mathbf  M$ such that $\mathbf M = \mathbf M_{ss}$ \cite{Huang_2023,fujiki2022series}.
It should be noted that when the static friction force of the system is less than the output force of the impedance control, the motion could be realized\cite{fujiki2022series}. Otherwise, it might not be possible to generate sufficient displacement as an input.
Then it becomes a force sensorless impedance controller as Eq.~\eqref{MB5}.
\begin{equation}
\label{MB5}
\begin{aligned}
\boldsymbol \tau_s = & \boldsymbol{\mathbf M_{ss}\ddot q_d + B(t)\dot q_d + K(t)q_d}\\
&  + \boldsymbol{(\mathbf C_{ss}-B(t))\dot q_s - K(t)q_s+\mathbf G_{s}}\\
&  + \boldsymbol{\mathbf M_{bs}^T\ddot q_b + \mathbf C_{sb}\dot q_b} - \boldsymbol{\tau_d}.
\end{aligned}
\end{equation}

The control input contains environmental interference item $\boldsymbol{\tau_d}$, which is not observable. 
To reduce the error of control input, we use VIC strategy to approximate this disturbance.
Let the impedance parameter update law be
\begin{equation}
\label{MB6}
\begin{aligned}
\boldsymbol{B(t)} = & \boldsymbol{B_0 + \Delta B},\\
\boldsymbol{K(t)} = & \boldsymbol{K_0 + \Delta K},
\end{aligned}
\end{equation}
then the control input would be written as
\begin{equation}
\label{MB7}
\begin{aligned}
\boldsymbol \tau_s = & \boldsymbol{\mathbf M_{ss}\ddot q_d + \mathbf B_0\dot q_d + \mathbf K_0q_d}\\
&  + \boldsymbol{(\mathbf C_{ss}-\mathbf B_0)\dot q_s - \mathbf K_0q_s+\mathbf G_{s}} + \boldsymbol{\mathbf M_{bs}^T\ddot q_b + \mathbf C_{sb}\dot q_b} \\
&  + \underbrace{\boldsymbol{\Delta B\boldsymbol{(\dot q_d-\dot q_s)} + \Delta K(q_d-q_s)}}_{\rm approximate\, disturbance},
\end{aligned}
\end{equation}
due to the external disturbance is unobservable, we use $\boldsymbol{\Delta B\boldsymbol{(\dot q_d-\dot q_s)} + \Delta K(q_d-q_s)}$ to replace $\boldsymbol{\tau_d}$. 
So it can satisfy $||\boldsymbol{\Delta B\boldsymbol{(\dot q_d-\dot q_s)} + \Delta K(q_d-q_s)}||_2<||\boldsymbol{\tau_d}||_2$ by setting impedance parameters variation. 
In the case of unobservable external environment disturbance, the VIC method can achieve more accurate control input.

\begin{figure}[tbp]    
\centering      
\includegraphics[width=1\columnwidth]{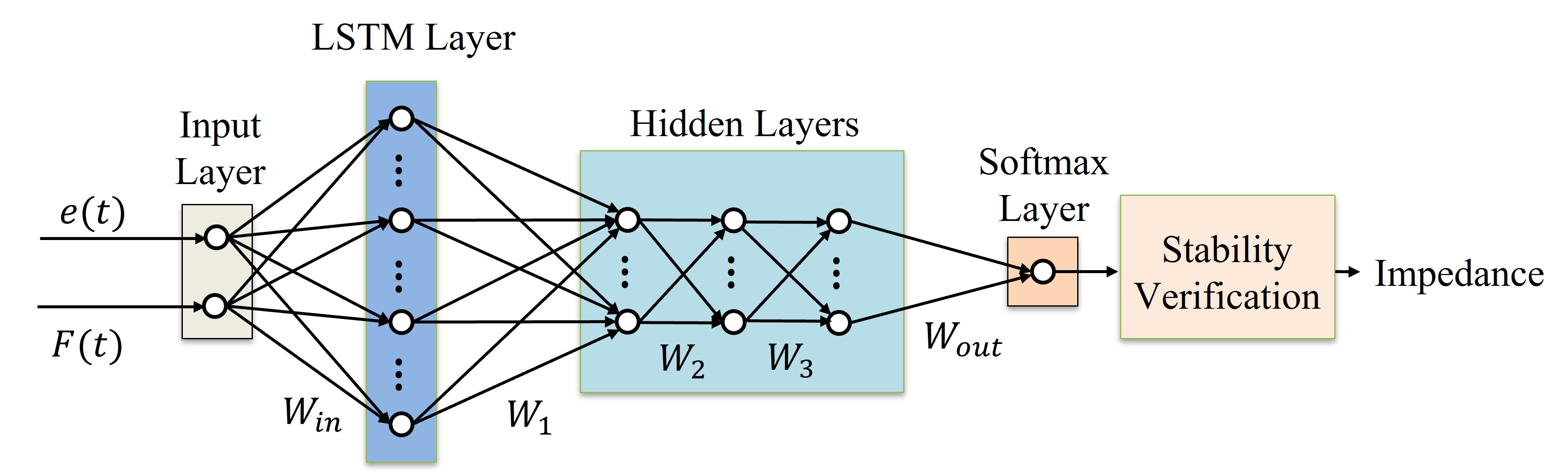}
\caption{RSG-IPGN for the VIC model.}  
\label{NNmodel}
\end{figure}

\subsection{RSG-IPGN for VIC Update Strategy}
\renewcommand\arraystretch{1.5}
\begin{table}[t]
    \caption{Tendency of desired impedance in each phase.}
    \label{Impe}
    \centering
        \begin{tabular}{cccc}
          \hline
          \hline
            Phase &  $e(t)$ & $F_e$ & Desired Impedance \\
            \hline
            Swing (SW) & Low  & Low & Low   \\  
            Contact Moment (CM) & High & $\nearrow$  &  Low   \\
            Stance (ST)  & High & High & High \\ 
           \hline
           \hline
        \end{tabular}
\end{table}

\begin{theorem}
The composite inequalities formula Eq.~(\ref{theo1}) guarantees closed-loop stability in the dynamics of the used VIC system, which is defined by Eq.~(\ref{MB2}).
\begin{equation}
\label{theo1}
\left\{
\begin{aligned}
&\alpha \boldsymbol B(t)+\boldsymbol K(t)-\alpha^2\mathbf M+\mathbf K_e \geq 0\\
&\boldsymbol B(t)-\alpha \mathbf M>0\\
&2\alpha \boldsymbol K(t)+ 2\alpha \mathbf K_e-\alpha \dot {\boldsymbol B}(t)-\dot {\boldsymbol K}(t)>0
\end{aligned}\right.
\end{equation}
The proof is given in supplementary material.
\end{theorem}

\begin{figure}[!tbp]    
\centering      
\includegraphics[width=1\columnwidth]{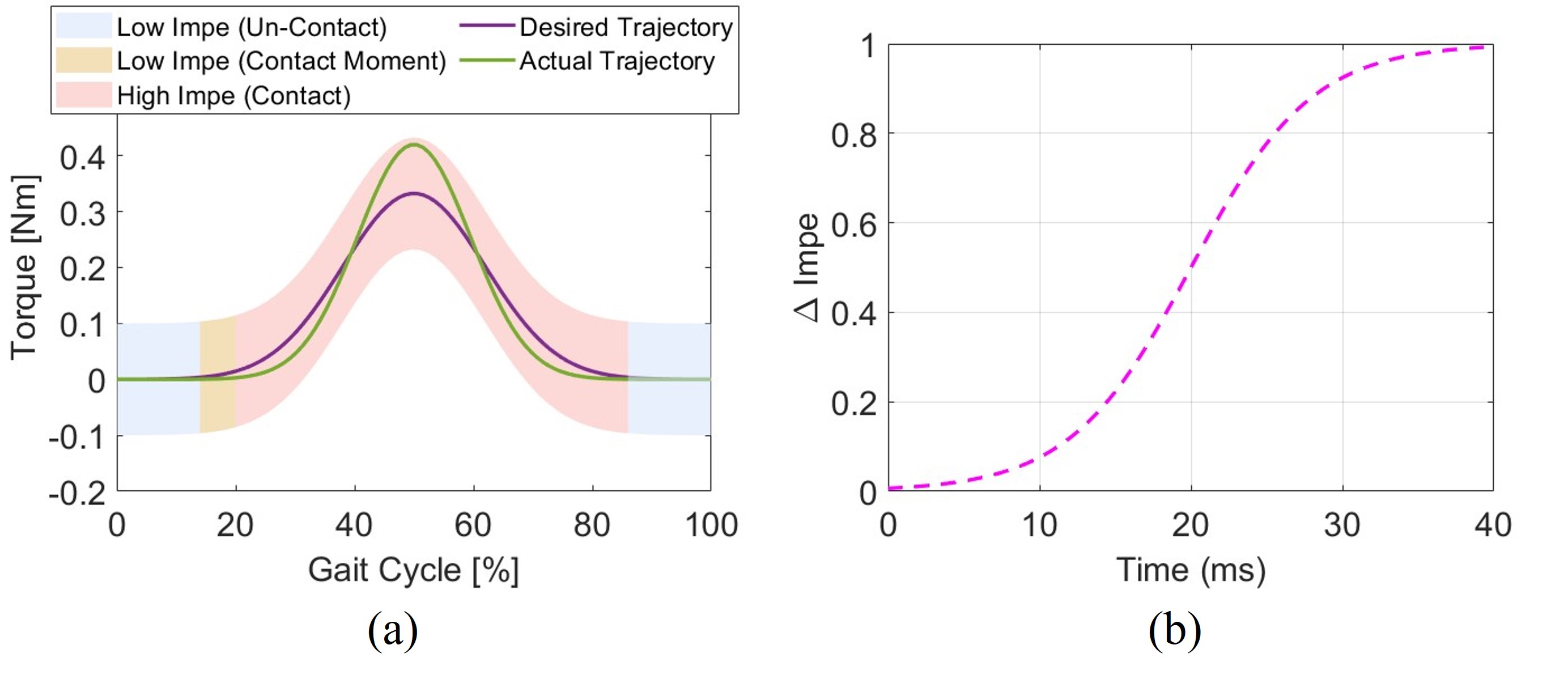}
\caption{(a) An example of schematic impedance illustration during different state. (b) Impedance transition smoothing function.}
\label{taskimpe}
\end{figure}

A real-time stability guaranteed network (shown in Fig.~\ref{NNmodel}) is applied to learn the time-varying impedance parameter update strategy. Let $\boldsymbol{e=q_s-q_d}$, Table ~\ref{Impe} shows the desired impedance of the task in different states. 
Fig.~\ref{taskimpe}(a) illustrated the state dependent impedance variation. 
When the mechanism is not in contact with the ground, i.e., during the swing phase, the system exhibits low impedance to avoid collisions, as shown in the blue area. Once the mechanism begins to make ground contact, the ground reaction force starts to increase, and the system continues to maintain low impedance while moving near the predetermined trajectory, as illustrated in the orange area. The purpose of maintaining low impedance during this phase is to absorb the impact forces upon ground contact, providing a cushioning and protective effect. In contrast, during the stance phase, where the system needs to provide support, it transitions to a high-impedance state, as shown in the red area.

\begin{figure*}[tbp]
\centering
\includegraphics[width=1\textwidth]{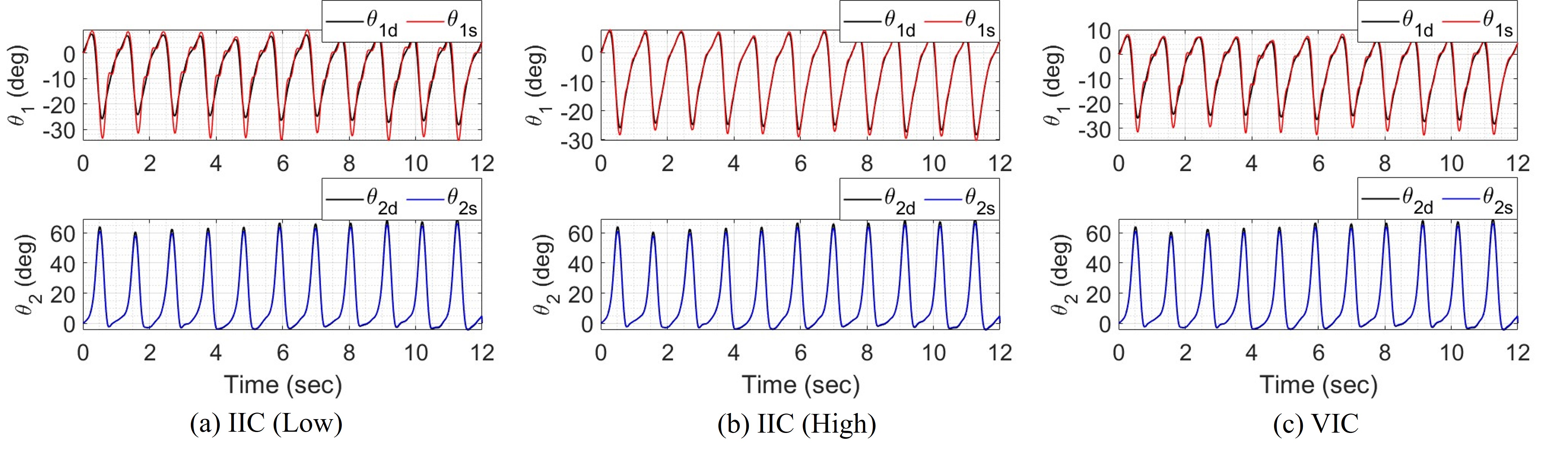}
\caption{Joint trajectory simulation results of SRL system. (a) IIC with low impedance controller. (b) IIC with high impedance controller. (c) VIC controller.}
\label{Simu_fig}
\end{figure*}

\begin{figure}[tbp]
\centering
\includegraphics[width=1\columnwidth]{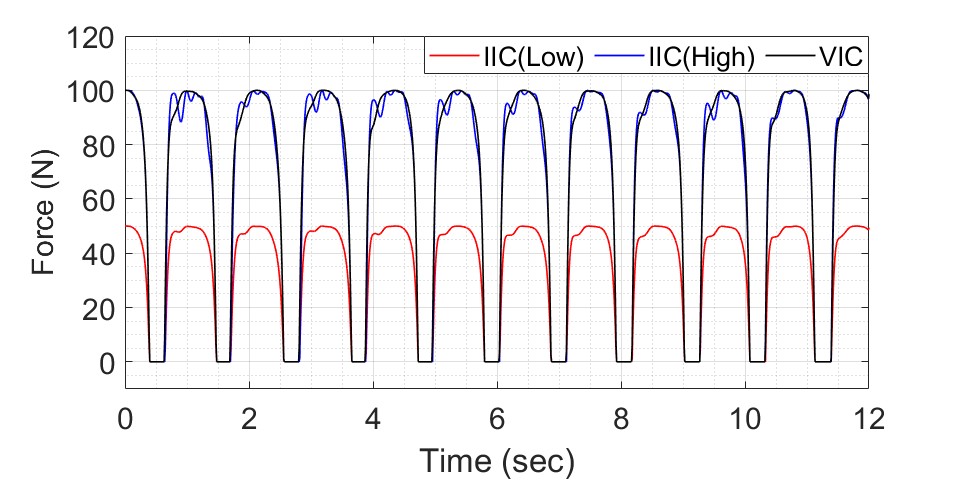}
\caption{Interaction force simulation results of the proposed SRL system.}
\label{Simu_fig2}
\end{figure}

First, the sensor data for different states is collected. After that, we built a neural network classifier model, calibrated the data according to the task impedance in the Table ~\ref{Impe}, and completed the offline network model pre-training.
During online impedance adjustment, the system classifies the current input to determine the desired impedance level in real time and outputs the corresponding class label. To prevent oscillations caused by frequent switching between high and low impedance states, a smoothing sigmoid function is incorporated, as illustrated in Fig.~\ref{taskimpe}(b).
The sigmoid function gain is
\begin{equation}
\label{SFun3}
s = \frac{1}{1+e^{(-a(x-b))}},
\end{equation}
where, $a=0.25$, $b=20$.
Then the impedance variation is designed as,
\begin{equation}
\label{SFun4}
\Delta I = s\cdot(\rm HI-LI),
\end{equation}
where, $\rm HI$ is the high impedance, $\rm LI$ is the low impedance.
In real-time system control, when the impedance parameter generator generates parameters that do not meet the stability conditions in the supplementary material, the system will skip this set of parameter values and use the impedance parameter of the previous cycle to control the system.

\section{Numerical Simulation Experiments}
In order to validate the ability of the proposed VIC framework, in this section, we performed simulations on the SRL system with gait data as system input. Physical parameters of the simulation study are displayed in supplementary material. 
The desired trajectory is collected by a subject walking in the treadmill.
A spring damping model is used to simulate the ground feedback force.
The simulation condition is Matlab/Simulink R2023a.

From the joint trajectory in Fig.~\ref{Simu_fig}, it can be observed that both IIC and VIC effectively track the predefined trajectory. Compared to the trajectory under IIC, the VIC trajectory exhibits a slight oscillation during the stance phase. This oscillation allows the SRL-assisted device to contact the ground earlier, gradually increase rigidity, and progressively apply support force, thereby providing an effectively upon ground contact. 
From the interaction force in Fig.~\ref{Simu_fig2}, at low impedance, the measured support force is low, while both high impedance and variable impedance can ensure greater support force. Compared with high impedance, the force signal of VIC is smoother during the peak rise phase.

The VIC strategy demonstrates consistent reduction in tracking error and stable high output force across varying conditions, confirming its feasibility and effectiveness. 
The effects of different controllers under varying input trajectories are examined through simulation, with the corresponding results presented in the supplementary material. It is found that the impact of trajectory variations on the experimental outcomes is weak.
Simulation provides a safe and efficient means for early validation, enabling controlled testing and statistical analysis without physical trials.

\begin{figure}[!t]    
\centering      
\includegraphics[width=1\columnwidth]{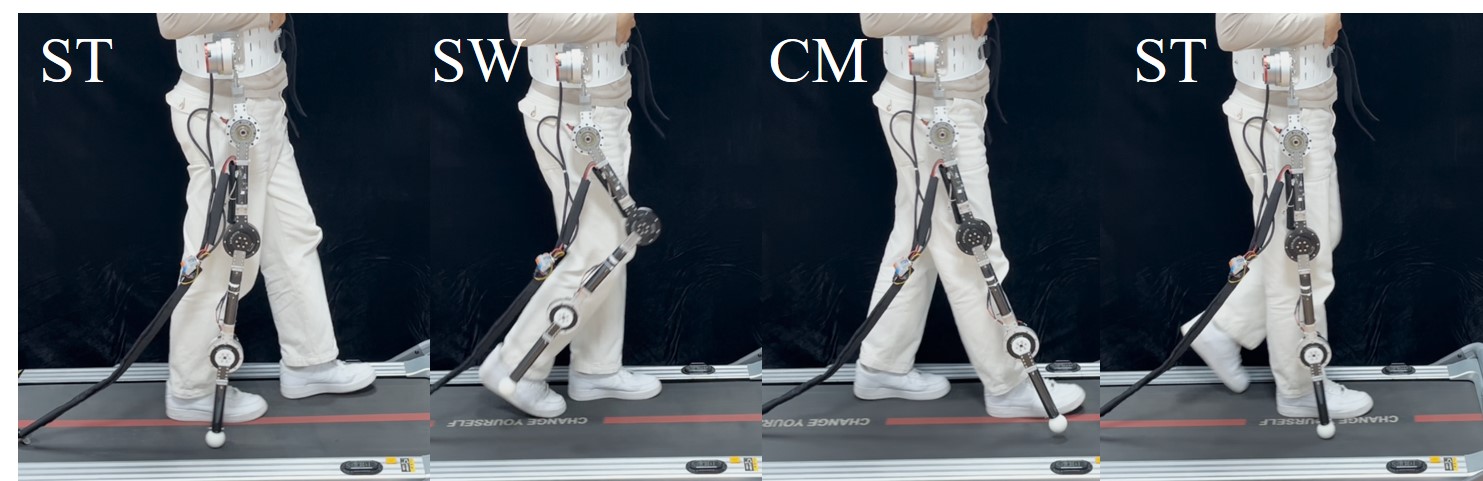}
\caption{A subject wearing the SRL device participates the experiments on the treadmill.}
\label{ExpSetup}
\end{figure}

\begin{figure*}[htbp]
\centering
\includegraphics[width=1\textwidth]{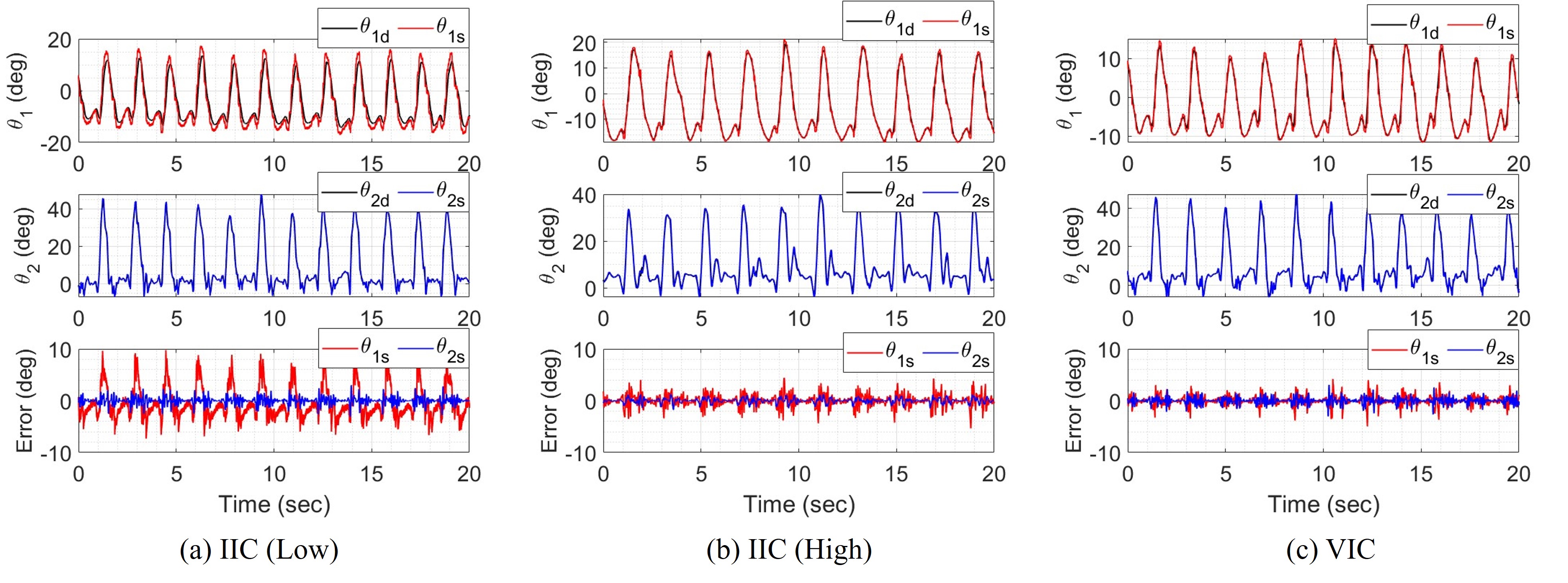}
\caption{Joint position trajectory of the experimental results. (a) IIC with low impedance controller. (b) IIC with high impedance controller. (c) VIC controller.}
\label{deg}
\end{figure*}

\begin{figure}[htbp]
\centering
\includegraphics[width=0.8\columnwidth]{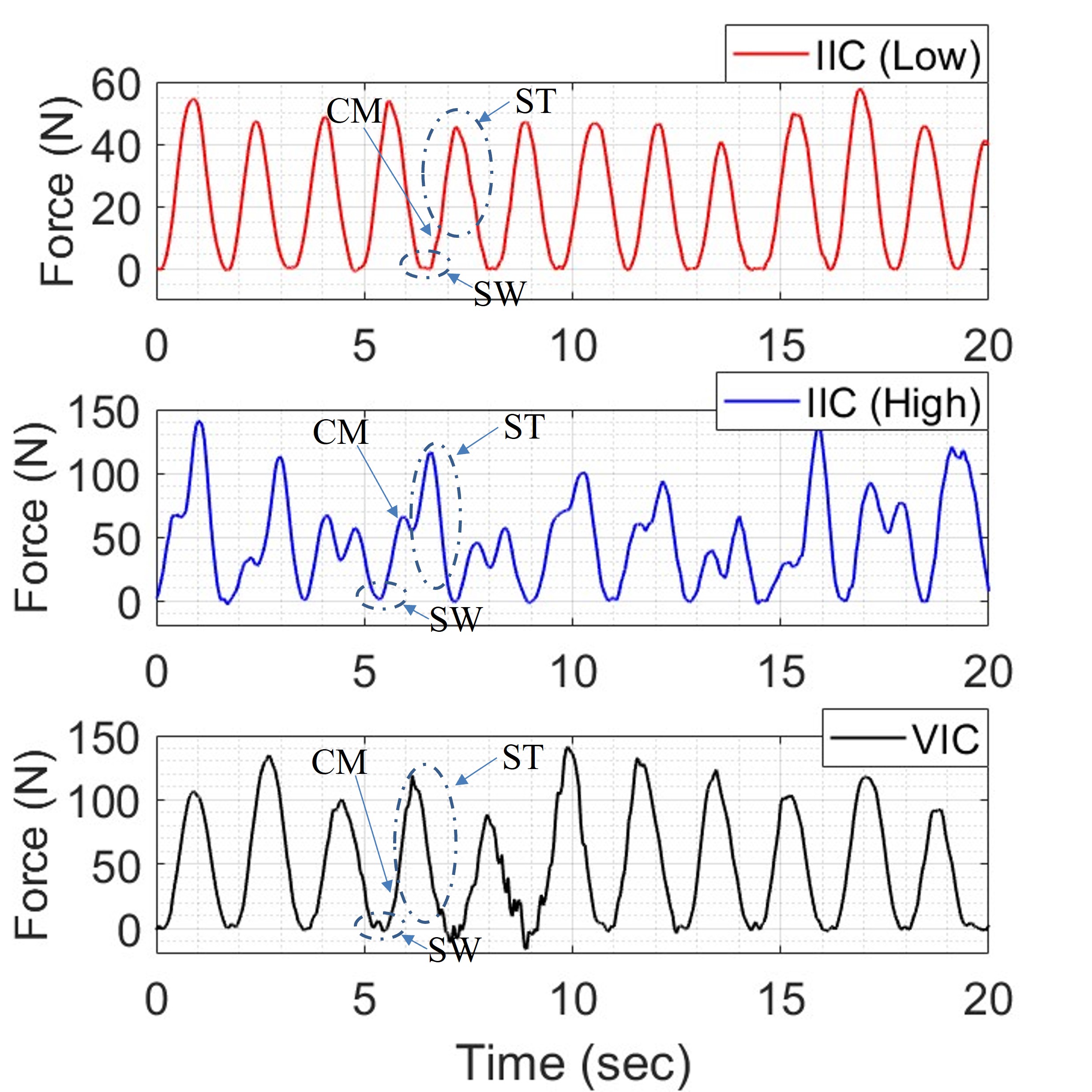}
\caption{Human-robot interaction force of the experimental results. }
\label{force}
\end{figure}
\section{Experiments and Discussions}
In this section, we conduct real human testing to demonstrate the efficacy of the proposed VIC in the SRL system.
\subsection{Implementation}
The specifications of main hardware devices are introduced in the supplementary material.
The experiments were conducted on a treadmill, as illustrated in Fig.~\ref{ExpSetup}, with a SRL device worn by the subject to aid with walking. 
Five healthy subjects participated in the experiment. The detailed information of the participants can be found in the supplementary material. The ethics approval for experiments with subjects was granted by the Ethics Committee of Tongji Medical College, Huazhong University of Science and Technology (NO. IORG0003571).

In our experiments with healthy participants, the SRL trajectory is generated based on the contralateral (left) leg to simulate the condition of a hemiplegic user with right-side impairment. Accordingly, the SRL is consistently mounted on the right side throughout all trials.
The desired trajectory $\boldsymbol q_d$ is generated by the method in Section II-A.

\subsection{Experimental Results}
The subject walked on the treadmill at speeds of 2.0 kph. In the experiment, the subject was asked to walk for 2 minutes each time, and the average value of 6 experiments was taken as the measurement result.
Root mean square jerk (RMSJ) metric is chosen as smoothness metric \cite{Balasubramanian2012}, which calculated as follows,
\begin{equation}
\label{RMSE}
{\rm RMSJ}=\sqrt{\frac{1}{t_2-t_1}\int_{t_1}^{t_2} \lvert\frac{d^2s}{dt^2}\rvert^2dt}.
\end{equation}

The joint angle and human-robot interaction force are shown as Fig.~\ref{deg} and Fig.~\ref{force}.
Under the low-impedance IIC controller, the SRL system exhibits high flexibility but is highly sensitive to external disturbances and fails to provide adequate support force (peak force is about 60 N in Fig.~\ref{force}), leading to trajectory deviations during ground contact (Fig.~\ref{deg}(a)). In contrast, the high-impedance IIC mode ensures greater support force but introduces undesirable impact forces due to oscillations at touchdown (an obvious fluctuation in CM of Fig.~\ref{force}). The VIC controller adaptively modulates impedance based on system state, maintaining low impedance at ground contact to absorb impact and high impedance during the support phase to enhance output force (Fig.~\ref{force}).

\begin{figure}[tbp]
  \centering
  \includegraphics[width=1\columnwidth]{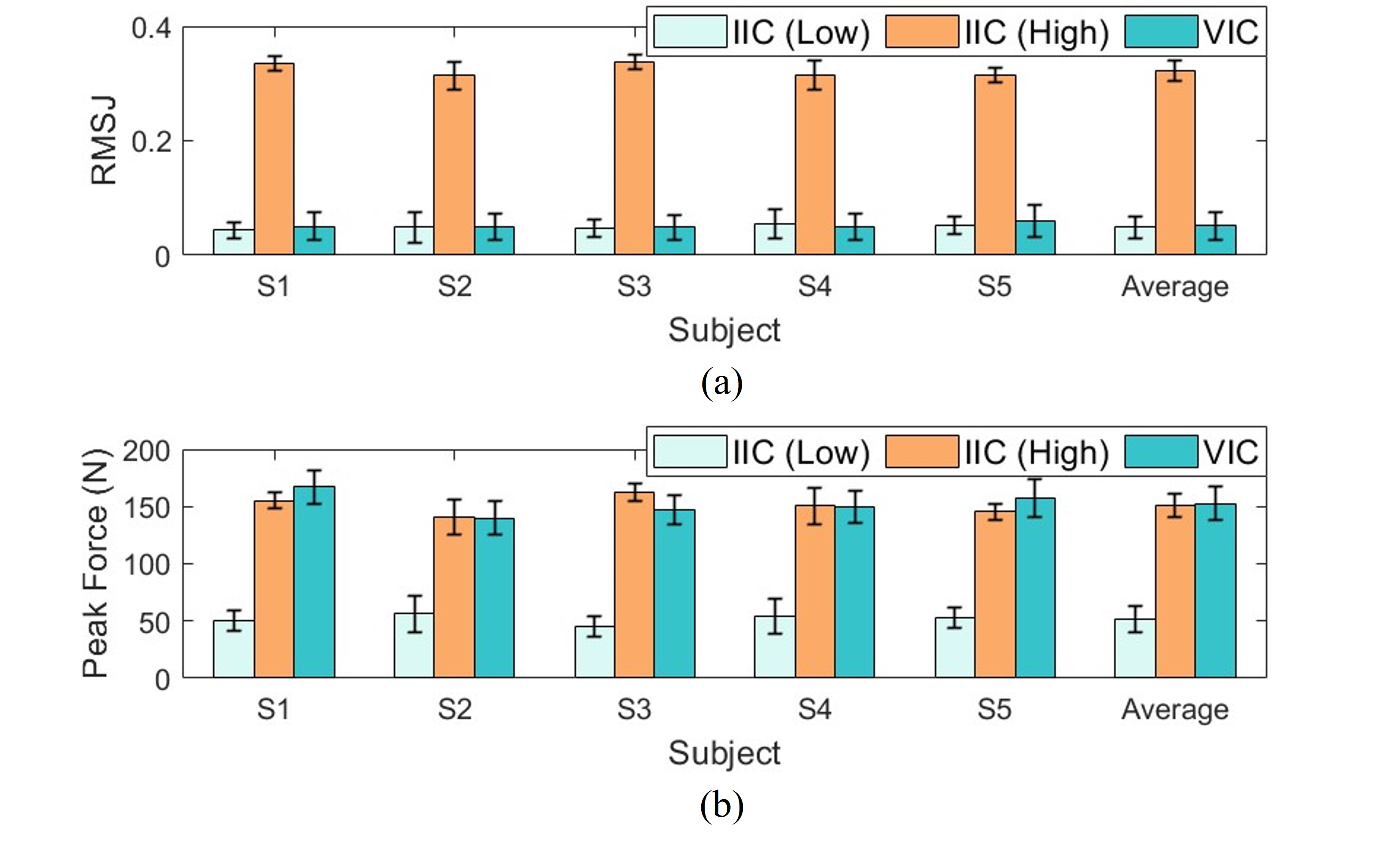}
  \caption{The quantitative comparison results of different subjects. (a) RMSJ index (b) Peak force index.}
  \label{ExpDisplay}
\end{figure}

Fig.~\ref{ExpDisplay} presents a comparison of experimental results between  VIC and IIC control modes. During the interactive force rising phase, the RMSJ index is used to assess force signal smoothness, with lower values indicating improved user comfort. The RMSJ is low under both low impedance and VIC modes, but significantly higher in the high impedance mode, suggesting the presence of impact forces at ground contact.
The peak force value reflects the maximum assistance provided. Both  high impedance and VIC modes deliver substantially higher peak forces than the low impedance mode, indicating their effectiveness in support during the stance phase.
To account for differences in participant height, the length of the  mechanism's connecting rod was adjusted accordingly. However, this variation had no significant effect on the statistical outcomes.
Overall, the VIC mode effectively integrates the flexibility of the low impedance mode during ground contact with the rigidity of the high impedance mode during support, offering a balanced and adaptive control strategy.

\subsection{Discussions}
The design of the SRL-assisted walking gait is inspired by the biomechanics of human crutch-aided walking. When using a crutch, the dominant or stronger hand is typically employed for support. In rehabilitation medicine, gait patterns are generally classified into three-point and two-point gait patterns.
In our study, the control objective for trajectory tracking is to simulate the trajectory of the user's healthy leg using the two-point method. From the perspective of enhancing limb function in healthy individuals, the contra-lateral leg should be selected to maintain balance.
In terms of rehabilitation training, the SRL should initially simulate the process of crutch-assisted walking, synchronizing its movement with the healthy leg. Furthermore, as the user gains walking ability, the SRL can be programmed to replicate the gait of the affected leg, providing partial support. This approach helps the user gradually adapt to a normal walking pattern.

The observed jerkiness and hesitation in the participant's gait may be attributed to a combination of factors, including the SRL's motion-derived from the contralateral leg-which was not fully synchronized with the ipsilateral natural leg, resulting in phase mismatches. This effect was more pronounced during early trials as participants adapted to the SRL's ground impacts. 
The participant was instructed to restrict arm movement, and the absence of arm swing on the SRL-attached side reflects typical hemiplegic gait, where the affected arm remains flexed. This setup reasonably approximates real-world conditions for the intended user population.
Given that natural arm swing plays a critical role in maintaining gait balance, angular momentum, and energy efficiency \cite{ORTEGA2008}, its suppression likely contributed to decreased gait stability and increased asymmetry. Furthermore, participants were instructed to walk conservatively for safety, which may have reinforced a cautious gait pattern. Notably, the pre-defined SRL trajectories were also generated under similar constraints, potentially affecting their naturalness. To account for variability across walking speeds, separate data collection sessions were conducted for each condition; the results are provided in the supplementary material and indicate no significant performance differences. In future work, we aim to cancel upper body constraints and optimize experimental ergonomics to  approximate natural locomotion better and improve the ecological validity of SRL-assisted gait evaluation.

Our control design is based on biomechanical function: hip joint uses variable impedance to handle ground reaction forces and ensure stability, while knee joint applies PID position control to maintain rigidity and track the swing trajectory. This also reflects biological patterns, where hip impedance varies more than knee impedance during gait \cite{Kooij}. Compared to dual-joint impedance control, our approach simplifies controller tuning and reduces dependence on precise model parameters, while maintaining adaptability and tracking accuracy.

Experiments showed that the SRL can reduce muscle activity in the opposite leg \cite{Takahito}. This effect supports various application scenarios: for individuals with hemiplegia, placing the SRL on the healthy side may help reduce fatigue in the affected leg, while positioning it on the affected side can facilitate rehabilitation training. 
Although our design was initially motivated by hemiplegic users-where the SRL compensates for an impaired limb-the anti-symmetric configuration is also applicable to unimpaired individuals.
For healthy users carrying a unilateral load, placing the SRL on the opposite side can help balance muscular effort between legs. These findings highlight the SRL's potential for personalized support in both daily life and therapeutic settings.
The anti-symmetric design inherently introduces asymmetry, which may cause discomfort or postural strain. Future work may address this through adaptive trajectory control to minimize lateral load shifts, and through mechanical design improvements such as adjustable stiffness or passive compliance at the interface.
Recent work \cite{abeywardena2024} demonstrates how simulation-based design from the biomechanics community  can assess SRL feasibility prior to prototyping. Integrating such predictive modeling tools (e.g., OpenSim) into the SRL design process would enable more systematic, human-centered development.

While the ultimate goal of SRL systems is to support individuals with  hemiplegia patients, healthy participants are recruited in this study to evaluate the baseline performance, control robustness, and safety of the proposed method under well-controlled conditions. Conducting experiments with patients at this stage poses significant safety risks and requires strict ethical review and approval. Therefore, preliminary evaluation with healthy individuals is a necessary and prudent step. This approach is consistent with standard practices in assistive robotics research and serves as a critical phase prior to clinical deployment.
In future work, we plan to conduct patient experiments in real-world rehabilitation scenarios.

\section{Conclusion}
This paper presents an anti-symmetric SRL to assist healthy individuals in walking tasks. As a typical wearable floating-base robot system, the control complexity is increased by both internal and external disturbances. To address internal disturbances, we propose a force-position hybrid impedance control strategy with an additional torso feedback term. For external environmental interference, a VIC strategy is introduced, leveraging a real-time stability-guaranteed impedance parameter generation network. The system’s impedance is dynamically updated, enabling it to switch between rigidity and flexibility depending on the phase of interaction.
The simulation and experiments on healthy people verified the excellent performance of the proposed VIC method.

\bibliographystyle{IEEEtran}
\bibliography{VICbib}



\end{document}